%% file: main.tex
\documentclass[runningheads]{llncs}

\usepackage{eccv}

\usepackage{eccvabbrv}

\usepackage{graphicx}
\usepackage{booktabs}
\usepackage{amsmath}
\usepackage{amssymb}

\usepackage[accsupp]{axessibility}  

\usepackage{hyperref}

\usepackage{orcidlink}

\begin{document}

\title{PANet: A Physics-guided Parametric Augmentation Net for Image Dehazing by Hazing} 

\titlerunning{PANet}

\author{Chih-Ling Chang\inst{1}$^\ast$ \and
Fu-Jen Tsai\inst{1}$^\ast$ \and 
Zi-Ling Huang\inst{2} \and Lin Gu\inst{2,3} 
\and \\ Chia-Wen Lin\inst{1}}

\authorrunning{C.-L. Chang et al.}

\institute{National Tsing Hua University \\ \email{jolene.clchang@gapp.nthu.edu.tw, fjtsai@gapp.nthu.edu.tw, cwlin@ee.nthu.edu.tw}\\ \and
The University of Tokyo, Japan \\
\email{huangziling@nii.ac.jp}\\
 \and
RIKEN, AIP\\
\email{lin.gu@riken.jp}}

\maketitle
\centerline{{$^\ast$ Equal Contribution.}}

\input{0Abstract}
\input{1Introduction}
\input{2RelatedWork}
\input{3ProposedMethod}

\input{4Experiments}
\input{5Conclusion}


%
%
\bibliographystyle{splncs04}
\bibliography{main}
\end{document}


\title{Supplementary Material \\
PANet: A Physics-guided Parametric Augmentation Net for Image Dehazing by Hazing} 

\titlerunning{Abbreviated paper title}

\author{First Author\inst{1}\orcidlink{0000-1111-2222-3333} \and
Second Author\inst{2,3}\orcidlink{1111-2222-3333-4444} \and
Third Author\inst{3}\orcidlink{2222--3333-4444-5555}}

\authorrunning{F.~Author et al.}

\institute{Princeton University, Princeton NJ 08544, USA \and
Springer Heidelberg, Tiergartenstr.~17, 69121 Heidelberg, Germany
\email{lncs@springer.com}\\
\url{http://www.springer.com/gp/computer-science/lncs} \and
ABC Institute, Rupert-Karls-University Heidelberg, Heidelberg, Germany\\
\email{\{abc,lncs\}@uni-heidelberg.de}}

\maketitle

In the supplementary material, we provide additional results to further validate the effectiveness of PANet. First, we give more performance evaluations between PANet and the GAN-based augmentation method: $D^4$~\cite{yang2022self}. Second, we demonstrate the effectiveness of PANet on real-world hazy scenes, which are captured without ground-truth reference images. Third, we demonstrate several augmented hazy images generated by PANet, along with their corresponding atmospheric light and density maps. Lastly, we discuss the limitations of PANet.      

\section{Comparison Between PANet and $D^4$}
\label{sec:efficiency}
In our method, we utilize a physics-guided learning strategy to optimize PANet, offering several advantages for PANet to generate realistic hazy images. First, PANet can map hazy images into a haze parameter space involving haze density and atmospheric light so that their haze conditions can easily adjusted in a physically meaningful manner by resampling the parameter space to generate diverse hazy images. Second, with the physical scattering model to generate initial hazy images as guidance, PANet can be realized with a relatively simple model that only contains $3$M parameters. Third, PANet does not rely on a large amount of training data. We can effectively optimize PANet using only $50$ pairs of hazy/clean images. 
In contrast, $D^4$ is on top of a CycleGAN-based architecture, which lacks the controllable ability to generate diverse hazy images. In addition, $D^4$ cannot be adequately optimized with only $50$ pairs of hazy/clean images since GAN-based methods require much more training data to learn robust statistic distributions. Furthermore, $D^4$ contains $11$M parameters, which is $8$M more than that of PANet.     

\section{Dehazing Results on Real-world Hazy Scenes}
To demonstrate the effectiveness of PANet in real-world hazy scenarios, we qualitatively compare the dehazing performances of dehazing models with or without using PANet-augmented data on the RTTS dataset~\cite{Resides}. Images in RTTS are collected in real-world hazy environments without ground truth reference images. We show the dehazed images with FocalNet~\cite{cui2023focal} in~\cref{fig:focal_rtts_1} and~\cref{fig:focal_rtts_2}, with DeHamer~\cite{guo2022dehamer} in~\cref{fig:dehamer_rtts_1} and~\cref{fig:dehamer_rtts_2}, and with DW-GAN~\cite{Fu_2021_CVPR} in~\cref{fig:dwgan_rtts_1} and~\cref{fig:dwgan_rtts_2}, where "Baseline" indicates the dehazing models without using PANet-augmented data, and "PANet" represents their PANet-enhanced versions.        

\section{Visuals of Hazy Images Augmented by PANet}
We then demonstrate several augmented hazy images by PANet and their corresponding haze density and atmospheric light maps in~\cref{fig:Augment_1} and~\cref{fig:Augment_2}. In the top parts of~\cref{fig:Augment_1} and~\cref{fig:Augment_2}, we show the original hazy images and their estimated haze density $\beta_\mathrm{est}(z)$ and atmospheric light $A_\mathrm{est}(z)$. In the bottom parts of~\cref{fig:Augment_1} and~\cref{fig:Augment_2}, we show the augmented hazy images and their corresponding haze density $\beta^{\prime}(z)$ and atmospheric light $A^{\prime}(z)$. By pixel-wisely resampling the hazy density and atmospheric light, we can generate diverse hazy images unseen in the training set. Since we parameterize various haze conditions into the haze density and atmospheric light to capture the characteristics of haze, the atmospheric light also contains haze-related information that can be used to adjust hazy patterns.  

\section{Limitations}
\label{sec:limitation}
The major limitation of PANet is that it relies on paired hazy/clean images to estimate pixel-wise haze parameters involving hazy density and atmospheric light. Nevertheless, PANet does not require tremendous amounts of training image pairs thanks to its simplified model based on the proposed physics guides, making it able to learn realistic haze distributions by only using much fewer training pairs. In addition, we believe that paired data are crucial for our current objectives since they enable us to learn accurate haze distributions from existing hazy images. Therefore, PANet offers an effective and efficient approach for augmenting real-world haze data and significantly improves dehazing models under real-world scenarios.

\begin{figure*}[t!]
\centering
\includegraphics[width=1\columnwidth]{24_Haze Augmentation/figure/focal_rtts_1.pdf}
\vspace{-0.3in}
\caption{Qualitative results of FocalNet~\cite{cui2023focal} on the RTTS~\cite{Resides} dataset.}
\label{fig:focal_rtts_1}
\vspace{-0.1in}
\end{figure*}

\begin{figure*}[t!]
\centering
\includegraphics[width=1\columnwidth]{24_Haze Augmentation/figure/focal_rtts_2.pdf}
\vspace{-0.3in}
\caption{Qualitative results of FocalNet~\cite{cui2023focal} on the RTTS~\cite{Resides} dataset.}
\label{fig:focal_rtts_2}
\vspace{-0.1in}
\end{figure*}

\begin{figure*}[t!]
\centering
\includegraphics[width=1\columnwidth]{24_Haze Augmentation/figure/dehamer_rtts_1.pdf}
\vspace{-0.3in}
\caption{Qualitative results of DeHamer~\cite{guo2022dehamer} on the RTTS~\cite{Resides} dataset.}
\label{fig:dehamer_rtts_1}
\vspace{-0.1in}
\end{figure*}

\begin{figure*}[t!]
\centering
\includegraphics[width=1\columnwidth]{24_Haze Augmentation/figure/dehamer_rtts_2.pdf}
\vspace{-0.3in}
\caption{Qualitative results of DeHamer~\cite{guo2022dehamer} on the RTTS~\cite{Resides} dataset.}
\label{fig:dehamer_rtts_2}
\vspace{-0.1in}
\end{figure*}

\begin{figure*}[t!]
\centering
\includegraphics[width=1\columnwidth]{24_Haze Augmentation/figure/dwgan_rtts_1.pdf}
\vspace{-0.3in}
\caption{Qualitative results of DW-GAN~\cite{Fu_2021_CVPR} on the RTTS~\cite{Resides} dataset.}
\label{fig:dwgan_rtts_1}
\vspace{-0.1in}
\end{figure*}

\begin{figure*}[t!]
\centering
\includegraphics[width=1\columnwidth]{24_Haze Augmentation/figure/dwgan_rtts_2.pdf}
\vspace{-0.3in}
\caption{Qualitative results of DW-GAN~\cite{Fu_2021_CVPR} on the RTTS~\cite{Resides} dataset.}
\label{fig:dwgan_rtts_2}
\vspace{-0.1in}
\end{figure*}

\begin{figure*}[t!]
\centering
\includegraphics[width=1\columnwidth]{24_Haze Augmentation/figure/augment_1.pdf}
\vspace{-0.3in}
\caption{Qualitative results of hazy images generated by PANet.}
\label{fig:Augment_1}
\vspace{-0.1in}
\end{figure*}

\begin{figure*}[t!]
\centering
\includegraphics[width=1\columnwidth]{24_Haze Augmentation/figure/augment_2.pdf}
\vspace{-0.3in}
\caption{Qualitative results of hazy images generated by PANet.}
\label{fig:Augment_2}
\vspace{-0.1in}
\end{figure*}


%
%
\bibliographystyle{splncs04}
\bibliography{main}

%% file: 0Abstract.tex
\begin{abstract}
 Image dehazing faces challenges when dealing with hazy images in real-world scenarios. A huge domain gap between synthetic and real-world haze images degrades dehazing performance in practical settings. However, collecting real-world image datasets for training dehazing models is challenging since both hazy and clean pairs must be captured under the same conditions. In this paper, we propose a Physics-guided Parametric Augmentation Network (PANet) that generates photo-realistic hazy and clean training pairs to effectively enhance real-world dehazing performance. PANet comprises a Haze-to-Parameter Mapper (HPM) to project hazy images into a parameter space and a Parameter-to-Haze Mapper (PHM) to map the resampled haze parameters back to hazy images. In the parameter space, we can pixel-wisely resample individual haze parameter maps to generate diverse hazy images with physically-explainable haze conditions unseen in the training set. Our experimental results demonstrate that PANet can augment diverse realistic hazy images to enrich existing hazy image benchmarks so as to effectively boost the performances of state-of-the-art image dehazing models.
  \keywords{Image Dehazing \and Haze Augmentation}
\end{abstract}

%% file: 1Introduction.tex
\section{Introduction}
\label{sec:intro}

Images taken in hazy environments often suffer from severe degradation, leading to undesirable contrast and appearance distortions. These hazy artifacts are often dense and non-homogeneous in real-world scenarios, significantly impacting the visual quality and visibility of scenes and downstream computer vision tasks, \eg object detection, tracking, and scene understanding. Image dehazing aims to restore high-quality and clean images from single hazy images, which is a highly ill-posed inverse problem due to information loss caused by the degradation.                

To address this problem, image dehazing techniques have found significant advancements through the success of deep learning. Numerous studies~\cite{9010659,Hardgan,cui2023focal,guo2022dehamer,song2023vision,FSDGN,li2019pdr,qu2019enhanced,AECR-Net} have been dedicated to improving dehazing performance through network architecture designs.  
These studies primarily utilized CNN-based modules to learn haze-related features, such as channel-wise attention~\cite{9010659}, haze-aware feature distillation~\cite{Hardgan}, and dual-domain selection module~\cite{cui2023focal}. 
In addition, inspired by the success of Transformers~\cite{vaswani2017attention} in vision applications~\cite{dosovitskiy2020vit,IPT,Ranftl21}, several studies have utilized Transformer-based architectures for image dehazing, such as transmission-aware Transformer~\cite{guo2022dehamer} and window-based Transformer~\cite{song2023vision}.
These methods primarily rely on synthetic image dehaze datasets~\cite{Resides} that adopt the physical scattering models~\cite{mccartney1976optics, nayar1999vision, narasimhan2003contrast} to generate homogeneous synthetic hazy images as 
\begin{equation}
\begin{gathered}
    I(z) = J(z)t(z)+A(1-t(z))
    \mbox{,} \\    
    t(z)=e^{-\beta d(z)}
    \mbox{,}
    \label{eq:physical scattering models}
\end{gathered}
\end{equation}
where $I(z)$ and $J(z)$ respectively denote the input hazy image and its corresponding clean version, $A$ and $t(z)$ denote the atmospheric light and transmission map, $\beta$ and $d(x)$ are the haze density and depth map, and $z$ is the pixel index.

\begin{figure}[t!]
    \begin{center}
    \includegraphics[width=1\textwidth]{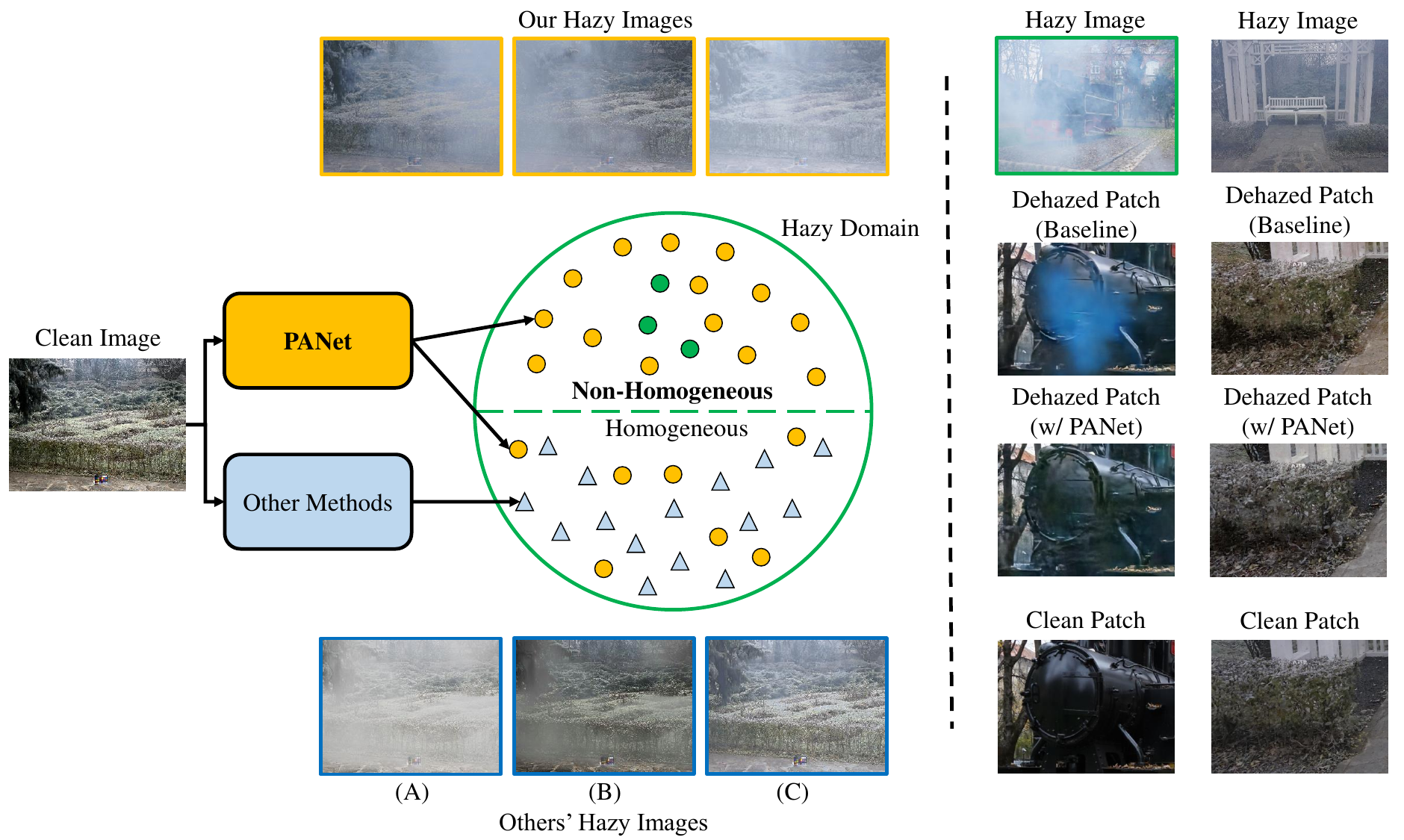}
    \end{center}
    \vspace{-0.2in}
    \caption{Left: Examples of the augmented images by PANet (yellow rectangles) and other augmentation methods (blue rectangles). Existing non-homogeneous hazy datasets~\cite{NH-Haze_2020, NH-Haze_2021} contain a very limited number of training pairs, as shown in green circles. Previous augmentation methods~\cite{wu2023ridcp,yang2022self} cannot effectively generate non-homogeneous hazy images. In contrast, PANet can generate realistic non-homogeneous and homogeneous hazy images, as shown in yellow rectangles. Right: Comparison of the dehazing results with and without using data augmented by PANet.} 
    \label{overview}
    \vspace{-0.1in}
\end{figure}

Although using the physical scattering models can generate abundant hazy and clean counterparts, a large domain gap between synthetic and real-world hazy image distributions usually significantly limits image dehazing performances in practical settings~\cite{gui2023comprehensive, zhang2021learning, Chen_2021_CVPR_PSD}, where real-world hazy images often exhibit dense and non-homogeneous haze~\cite{Dense-Haze_2019,NH-Haze_2020,NH-Haze_2021}, as shown in Fig.~\ref{overview}. 
To address this problem, existing works~\cite{NH-Haze_2020,NH-Haze_2021} mainly resort to collecting real-world non-homogeneous hazy and clean training image pairs. However, collecting such datasets is challenging and costly since both the hazy and clean counterparts must be captured under the same conditions, including identical moving objects and background light. Therefore, these datasets often contain a very limited number of training pairs, significantly limiting the performances of deep dehazing models in real-world scenarios.
Some methods have attempted to enhance the diversity of hazy images via brightness adjustments~\cite{wu2023ridcp} or global haze density adjustments~\cite{yang2022self}, as illustrated in Figs.~\ref{overview}(B) and \ref{overview}(C). Nevertheless, they ignore the above important fact that real-world haze distributions are often dense and non-homogeneous, making the domain gap still large. 
Therefore, it is crucial to propose a new approach to learn to generate additional realistic non-homogeneous hazy images with various haze conditions from existing hazy and clean image pairs without heavily relying on a high-cost data collection process.  

\begin{figure}[t!]
    \begin{center}
    \includegraphics[width=1\textwidth]{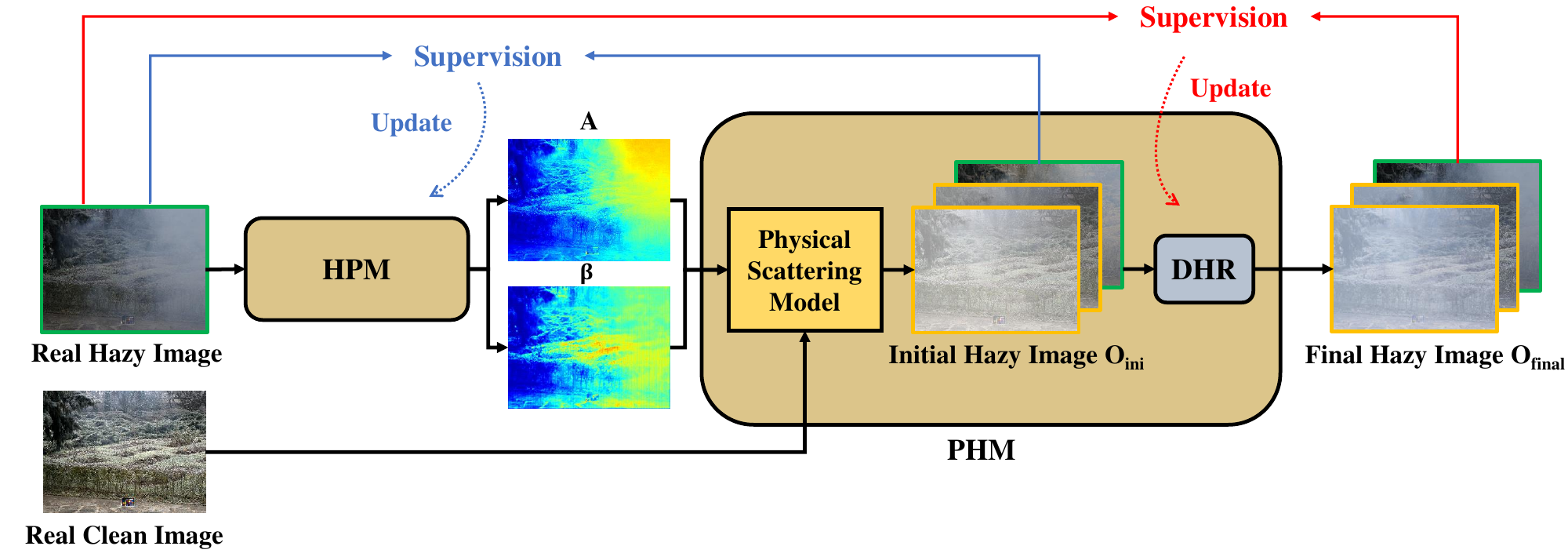}
    \end{center}
    \vspace{-0.2in}
    \caption{Overview of the proposed PANet. PANet comprises a Haze-to-Parameter Mapper (HPM) to project real haze images into a parameter space and a Parameter-to-Haze Mapper (PHM) to revert them back to the real space. In addition to generating original hazy images (green rectangle box), PANet can generate real hazy images not provided in the training set (yellow rectangle box).} 
    \label{fig2}
    \vspace{-0.3in}
\end{figure} 

In this paper, we propose a novel Physics-guided Parametric Haze Augmentation Net (PANet) to effectively augment realistic non-homogeneous hazy and clean training pairs for enhancing dehazing performance in the real world. As overviewed in Fig.~\ref{fig2}, PANet utilizes a Haze-to-Parameter Mapper (HPM) to project the hazy images of training clean and hazy image pairs into a parameter space spanned by haze parameters, followed by a Parameter-to-Haze Mapper (PHM) to resample and then convert the haze parameters in the parameter space back to haze images. Specifically, inspired by the widely-adopted Physical Scattering Model (PSM) in Eq. (\ref{eq:physical scattering models}), HPM parameterizes the haze conditions of hazy images into two pixel-wise haze parameters: haze density and atmospheric light. Next, these estimated parameters are resampled and then used to translate the clean version into its hazy versions with various haze conditions in a two-step manner. In the initial step, we utilize the physical scattering model in Eq. (\ref{eq:physical scattering models}) to generate reasonable initial hazy images. In the refinement step, we further use a Data-driven Haze Refiner (DHR) $N_\mathrm{DHR}(\cdot)$ to refine the initial hazy image to make it realistic. Note, although with the physics guides, since a scene in a dense and non-homogeneous haze is usually substantially occluded or distorted, it is still a highly ill-posed problem to retrieve accurate pixel-wise haze parameter maps from a clean and hazy image pair. Such ill-posedness makes the estimated haze parameters inaccurate, thereby leading to unsatisfactory visual qualities of hazy images generated solely by the PSM~\cite{gui2023comprehensive, zhang2021learning, Chen_2021_CVPR_PSD}. To address this problem, we propose using the DHR to refine the the input hazy image and then use the reconstruction error as supervision to form a cyclic haze-parameter-haze learning path involving both the HPM and DHR as elaborated in Sec. 3.  

Our proposed physics-guided haze augmentation scheme offers several advantages. First, with the scattering model-based physics guides, the estimated haze parameters have their physical meanings, making the parameter resampling and the resulting hazy image generation explainable. Second, our haze-parameter-haze mapping framework constitutes a cyclic learning path for haze generation involving haze parameter estimation,  model-based initialization, and data-driven refinement. This leads to more accurate haze parameter estimation and realistic hazy image generation to complement the inaccuracy of model-based generation as well as simply simplify the pure data-driven deep model design thanks to reasonable initialization based on the scattering models. Third, by resampling the explainable haze parameters, we can easily augment hazy images with various haze conditions to significantly boost the performances of existing deep dehazing models while reducing the cost of training data collection.
%
%
%
The contributions of this work can be summarized as follows:

\begin{itemize}
    \item We propose a novel physics-guided Parametric Augmentation Net (PANet) that generates realistic, non-homogeneous hazy images to enhance existing deep dehazing models.
    \item The proposed PANet is a controllable network that can retrieve from input hazy images core haze parameters, including pixel-wise haze density and atmospheric light,  and then resample the haze parameters to generate additional photo-realistic hazy images with various explainable haze conditions unseen in the original training set.
    \item Extensive experimental results demonstrate the high efficacy of PANet in boosting state-of-the-art dehazing models on four real-world image dehazing datasets. Cross-dataset evaluations also validate the generability of PANet.
\end{itemize}

%% file: 2RelatedWork.tex
\section{Related Work}
\subsection{Image Dehazing}
Image dehazing techniques have achieved remarkable progress with the fast growth of CNNs. Specifically, to effectively extract haze-related features, several studies resorted to attention-based methods using CNNs. For example, Liu~\etal~\cite{9010659} proposed a multi-scale attention-based network that utilizes channel-wise attention for feature fusion. Xu~\etal~\cite{qin2020ffa} proposed a feature fusion attention network with cascaded channel-attention and pixel-attention modules. Deng~\etal~\cite{Hardgan} proposed a haze-aware representation distillation module to distill haze-related features through instance normalization. Fu~\etal~\cite{Fu_2021_CVPR} utilized discrete wavelet transform with a generative adversarial network (GAN) to preserve high-frequency knowledge in the feature space. Cui~\etal~\cite{cui2023focal} proposed an efficient image restoration network that contains a dual-domain selection mechanism to emphasize important regions for restoration.

Recently, motivated by Transformers' powerful ability to model long-range dependencies among features, several studies have devised transformer-based models for image dehazing. Song~\etal~\cite{song2023vision} utilized window-based attention~\cite{liu2021Swin} to design a vision transformer for image dehazing. Guo~\etal~\cite{guo2022dehamer} proposed a hybrid architecture that integrates CNN and Transformer with a transmission-aware 3D position embedding to improve image dehazing performance. 
Although these methods successfully improve image dehazing performance through elaborate model designs, they primarily rely on synthetic hazy datasets, which may lead to a performance decrease when handling real-world hazy images. Instead of concentrating solely on architectural designs to improve dehazing performance, our goal is to design a haze augmentation method applicable across various dehazing models and improve dehazing performance in real-world scenarios. 

\subsection{Hazy Image Augmentation}
In addition to improving dehazing performance through architectural designs, some studies focused on hazy image augmentation strategies to enhance dehazing performance. Based on a physical scattering model, the method proposed in~\cite{wu2023ridcp} incorporates brightness adjustment, color bias, and Gaussian noise within the physical scattering model to simulate adverse light conditions in real-world scenarios. However, this approach modifies additional factors rather than leveraging the inductive bias of real-world hazy images, which are usually non-homogeneous with high opacity. 
To generate diverse hazy images with real-world haze characteristics, Yang~\etal~\cite{yang2022self} proposed a rehazing model that incorporates depth and haze density with CycleGAN~\cite{CycleGAN2017}. By globally sampling haze density, they can generate additional hazy images as a data augmentation operation. However, GAN-based architectures often encounter challenges such as unstable training process~\cite{NIPS2017_892c3b1c,8237566}, model collapse~\cite{asveegan17,MSGAN}, and uncontrollable outputs~\cite{KowalskiECCV2020,Shoshan_2021_ICCV}, which restricts the diversity and usability of the generated images. Furthermore, their method only allows global haze density adjustment, making it unsuitable for real-world hazy images that typically exhibit non-uniformity with high opacity. 

In contrast, we propose PANet, a stable and controllable framework to generate real-world hazy images for data augmentation, leveraging the inductive bias inherent in real-world hazy images. In addition, our approach allows for pixel-wisely adjusting haze conditions to generate non-homogeneous haze with different densities and positions. This enhances the diversity of hazy images and leads to
significant improvements on existing image dehazing models on several real-world hazy image datasets.

%% file: 3ProposedMethod.tex
\section{Proposed Method}
\subsection{Overview}
\begin{figure}[t!]
    \begin{center}
    \includegraphics[width=1\textwidth]{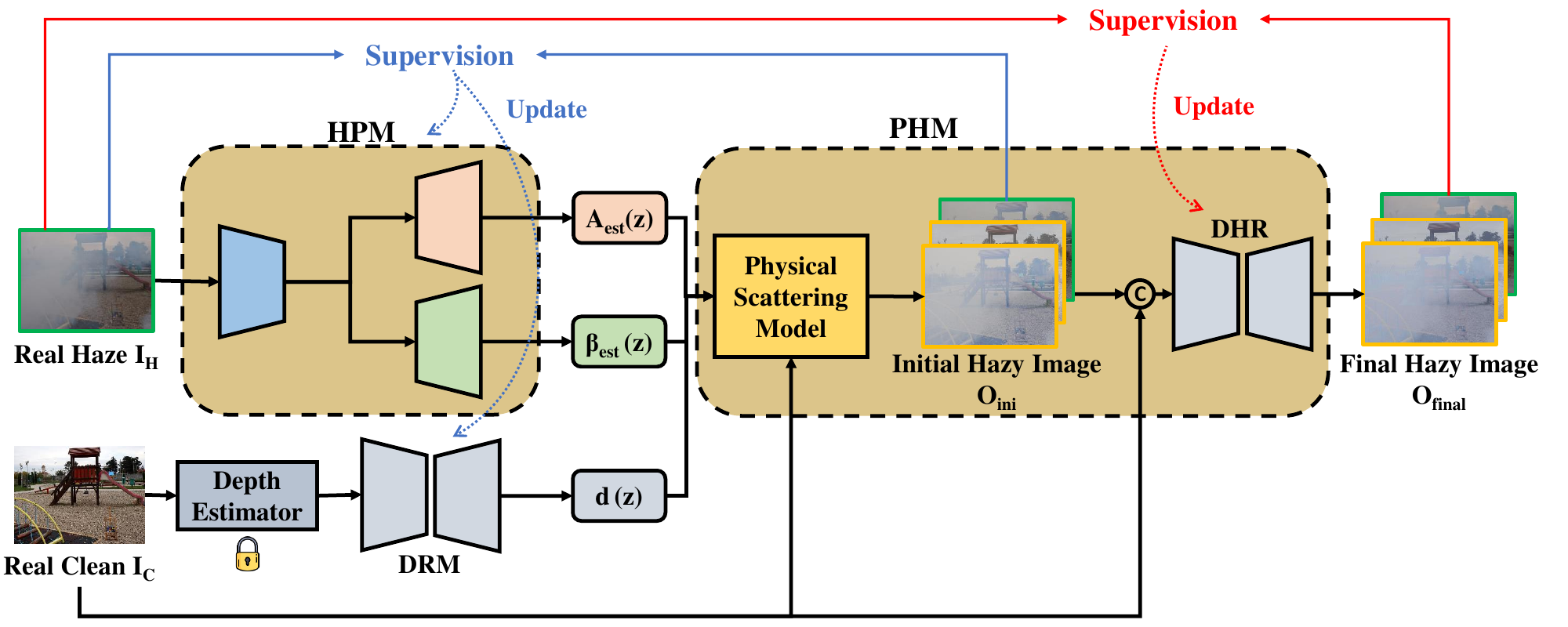}
    \end{center}
    \vspace{-0.2in}
    \caption{Block diagram of PANet. PANet utilizes a cyclic haze-parameter-haze mapping framework consisting of a Haze-to-Parameter Mapper (HPM) followed by a Parameter-to-Haze Mapper (PHM). Besides the hazy images in the original training set (green boxes), PANet can augment additional hazy images with various haze conditions unseen in the training set (yellow boxes).} 
    \label{fig3}
    \vspace{-0.2in}
\end{figure}

In the real world, haze is often non-homogeneous and exhibits high opacity. To generate hazy images with real-world characteristics, we propose a PANet that augments photo-realistic hazy images with diverse haze conditions for individual hazy and clean training pairs to effectively boost the performances of real-world image dehazing. 
Fig.~\ref{fig3} illustrates the block diagram of PANet, a cyclic Haze-Parameter-Haze mapping framework consisting of a Haze-to-Parameter Mapper (HPM) followed by a Parameter-to-Haze Mapper (PHM). Given a pair of hazy and clean images, we first use HPM to project the hazy image into a learned parameter space that characterizes real-world haze conditions with two physically-explainable haze parameters: the haze density and atmospheric light. Next, we use the estimated haze density and atmospheric light maps to translate the input clean image to an initial hazy image based on the physical scattering model in Eq. (\ref{eq:physical scattering models}). We then refine the initial hazy image to a more photo-realistic one by using a Data-driven Haze Refiner (DHR) $N_\mathrm{DHR}(\cdot)$ to complement the inaccuracy of the scattering model due to the ill-posedness of haze parameter estimation. 

In this way, besides the hazy images in the original training set, we can easily augment additional hazy iamges with various explainable haze conditions unseen in the training set by resampling the pixel-wise parametric haze conditions (\ie, the haze density and atmosphere light). This allows us to significantly enrich the training data, thereby enhancing the performances of existing dehazing models in real-world scenarios. Next, we will introduce the core components of PANet, including HPM, PHM, and the haze augmentation process. 

\subsection{Haze-to-Parameter Mapper (HPM)}
\begin{figure}[t!]
    \begin{center}
    \includegraphics[width=0.9\textwidth]{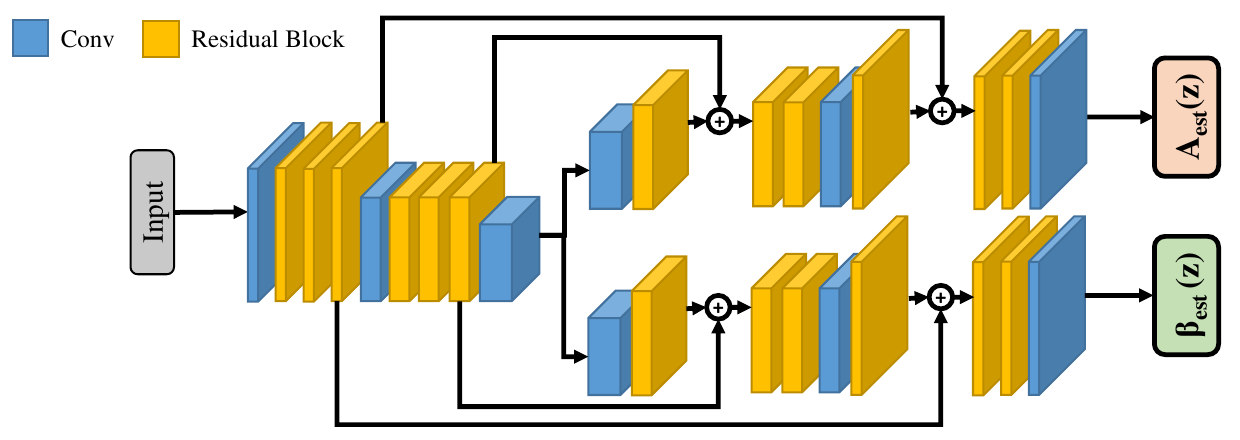}
    \end{center}
    \vspace{-0.2in}
    \caption{Architecture of Haze-to-Parameter Mapper (HPM). HPM consists of a shared encoder followed by two parallel parameter decoders to estimate the haze density map $\beta_\mathrm{est}(z)$ and atmospheric light map $A_\mathrm{est}(z)$, respectively.} 
    \label{encoderdecoder}
    \vspace{-0.2in}
\end{figure}

HPM aims to project hazy images into a learned parameter space that parameterizes real-world haze conditions with two physically-explainable haze parameters: haze density and atmospheric light. 
HPM comprises an encoder and two decoders, as shown in Fig.~\ref{encoderdecoder}. Given a hazy image $I_H(z) \in \mathbb{R}^{H \times W \times 3}$, where $z$ is the pixel index, the encoder $E_\mathrm{haze}(\cdot)$ extracts the haze-relevant features of $I_H(z)$. Next, the Density Decoder $D_\mathrm{dense}(\cdot)$ and the  
Atmospheric Light Decoder $D_\mathrm{AL}(\cdot)$ are then used to estimate the pixel-wise haze density map $\beta_\mathrm{est}(z) \in \mathbb{R}^{H \times W \times 3}$ and atmospheric light map $A_\mathrm{est}(z) \in \mathbb{R}^{H \times W \times 1}$, respectively, as 
\begin{equation}
\begin{gathered}
    \beta_\mathrm{est}(z) = D_\mathrm{dense}(E_{\mathrm{haze}}(I_H(z))),
    \label{eq:density}
\end{gathered}
\end{equation}
and
\begin{equation}
\begin{gathered}
    A_\mathrm{est}(z) =  D_\mathrm{AL}(E_{\mathrm{haze}}(I_H(z))). 
    \label{eq:atmosphere}
\end{gathered}
\end{equation}

To derive $d(z)$ in Eq. (\ref{eq:physical scattering models}), we choose RA-Depth~\cite{he_ra_depth} as the pre-trained depth estimator $\Psi(\cdot)$ same as RIDCP~\cite{wu2023ridcp}. Besides, to bridge the domain gap with the pre-trained depth estimator, we further use a Depth Refinement Module (DRM) $d_\mathrm{ref}$ to refine the depth map as 
\begin{equation}
    d(z) = d_\mathrm{ref}(\Psi(I_C(z))),
    \label{eq:d_predictor}
\end{equation}
where the architecture of $d_\mathrm{ref}(\cdot)$ is similar to HPM but with only one decoder. 

To train the modules in HPM to learn to accurately estimate $\beta_\mathrm{est}(z)$, $A_\mathrm{est}(z)$ and $d(z)$, we calculate the fidelity between the input hazy image and its reconstructed version generated from its clean version by using the PSM in Eq. (\ref{eq:physical scattering models}) and the PHM in Sec. \ref{sec: PHM} based on the predicted $\beta_\mathrm{est}(z)$, $A_\mathrm{est}(z)$, and $d(z)$. This reconstruction fidelity reflects the accuracy of the estimated parameters, thereby allowing us to use it as supervision to guide the learning of HPM and DRM.

\subsection{Parameter-to-Haze Mapper (PHM)}
\label{sec: PHM}
After retrieving the haze parameters and scene depth, we further utilize PHM to map the haze parameters in the parameter space back to real hazy images. Specifically, based on the estimated $\beta_\mathrm{est}(z)$ and $A_\mathrm{est}(z)$, we subsequently translate the input clean image $I_C(z)$ to the initial hazy image $O_\mathrm{ini}(z)\in \mathbb{R}^{H \times W \times 3}$ using the following Physical Scattering Model:
\begin{equation}
\begin{gathered}
    O_\mathrm{ini}(z) = I_C(z)t(z)+A_\mathrm{est}(z)(1-t(z)),     
    \label{eq:coarse hazy image}
\end{gathered}
\end{equation}
where
\begin{equation}
\begin{gathered}  
    t(z)=e^{-\beta_\mathrm{est}(z) d(z)},
    \label{eq:coarse hazy image_transmission}
\end{gathered}
\end{equation}
where $d(z)\in \mathbb{R}^{H \times W \times 1}$ denotes the depth map estimated from the clean image $I_C(z)$. 

By using the Physical Scattering Model, we can generate initial hazy images $O_\mathrm{ini}(z)$ from the clean image $I_C(z)$. This model provides physical meanings for the haze parameter estimated by HPM.
However, since a scene in a dense, non-homogeneous haze is usually substantially occluded or distorted, retrieving pixel-wise haze parameters from a hazy image is highly ill-posed, making the model-generated initial hazy images inaccurate and distorted (\eg, incorrect color tone and unrealistic transparency)~\cite{gui2023comprehensive, zhang2021learning, Chen_2021_CVPR_PSD}. Therefore, we propose a Data-driven Haze Refiner (DHR) $N_\mathrm{DHR}(\cdot)$ to further refine the initial hazy images to mitigate the inaccuracy.
To this end, we concatenate $O_\mathrm{ini}(z)$ with its corresponding clean image $I_C(z)$ and feed them to $N_\mathrm{DHR}(\cdot)$ to get the real hazy image $O_\mathrm{final}(z)$ as 
\begin{equation}
    O_\mathrm{final}(z) = N_\mathrm{DHR}(\mathbf{Concate}(O_\mathrm{ini}(z), I_C(z))),
    \label{eq:PHM}
\end{equation}
where  $N_\mathrm{DHR}(\cdot)$ has a similar architecture to HPM but with only one decoder.

With the cyclic mapping framework involving HPM and PHM, PANet can successfully project hazy images into a parameter space and then generate additional hazy images by pixel-wisely resampling the haze density $\beta_\mathrm{est}(z)$ and atmospheric light $A_\mathrm{est}(z)$ with physically-explainable haze conditions unseen in the training set, as elaborated in the haze augmentation process.    

\subsection{Loss Function}
We choose Charbonnier loss~\cite{Lai_2017_CVPR} $\mathcal{L}_\mathrm{char}$ and perceptual loss~\cite{perceptual} $\mathcal{L}_\mathrm{perc}$ for optimizing PANet as follows:
\begin{equation}
\begin{gathered}
\mathcal{L}_{\mathrm{total}}=\mathcal{L}_\mathrm{char}(O_\mathrm{ini}(z), I_H(z)) +\mathcal{L}_\mathrm{char}(O_\mathrm{final}(z), I_H(z)) \\
+ \lambda \mathcal{L}_\mathrm{perc}(O_\mathrm{ini}(z), I_H(z)) + \lambda \mathcal{L}_\mathrm{perc}(O_\mathrm{final}(z), I_H(z))
\mbox{,}
\label{eq:Char_loss}   
\end{gathered}
\end{equation}
where $O_\mathrm{ini}(z)$ denotes the model-generated initial hazy image, $O_\mathrm{final}(z)$ denotes the refined hazy images, $I_H(z)$ denotes the ground-truth hazy image, and $\lambda$ is a weight empirically set to $\lambda=10^{-6}$.

\begin{figure}[t!]
    \begin{center}
    \includegraphics[width=1\textwidth]{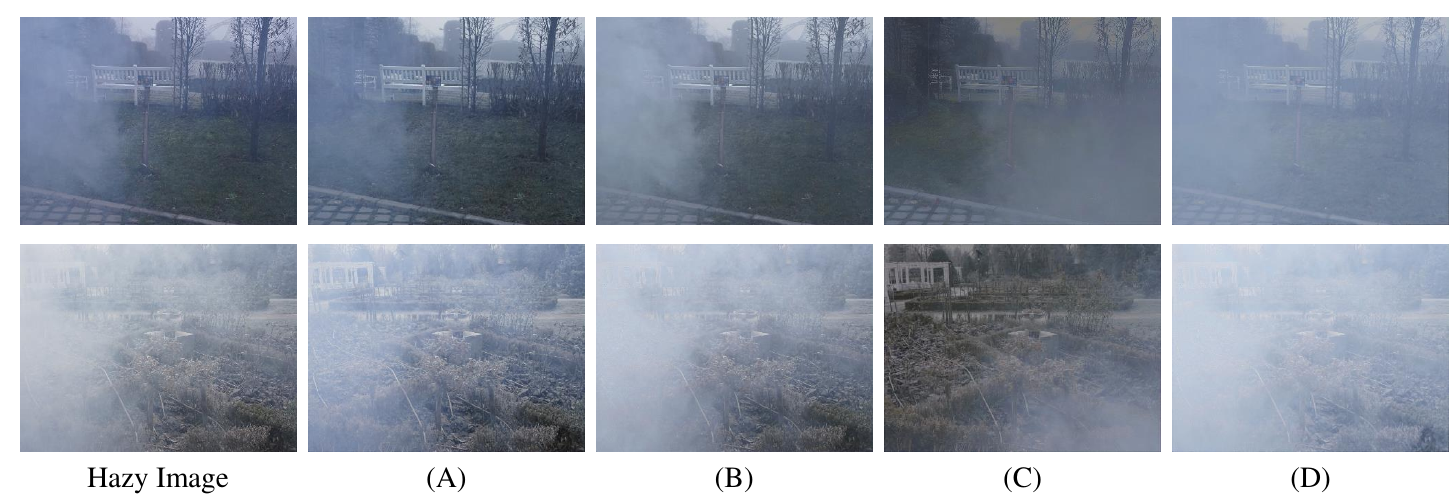}
    \end{center}
    \vspace{-0.2in}
    \caption{Visuals of hazy images generated by PANet. Given a hazy image, we can decrease or amplify its haze density by 2, as shown in (A) and (B). In addition, we can reverse its haze location or generate a complex hazy image, as shown in (C) and (D)} 
    \label{haze augmentation}
    \vspace{-0.2in}
\end{figure} 

\subsection{Haze Augmentation Process}
To generate new hazy images unseen in the training set, given a pair of hazy $I_H(z)$ and clean $I_C(z)$ images, we pixel-wisely resample its estimated haze density map $\beta_\mathrm{est}(z)\in \mathbb{R}^{H \times W \times 3}$ and atmospheric light map $A_\mathrm{est}(z)\in \mathbb{R}^{H \times W \times 1}$ to obtain their new versions: $\beta^{\prime}(z)$ and  $A^{\prime}(z)$. The two new maps are then used to generate a new hazy image by using PHM. Specifically, we can alter haze density $\beta_\mathrm{est}(z)$ by a scaling factor $\alpha$ to generate $\beta^{\prime}(z)$ as    
\begin{equation}
    \beta^{\prime}(z) = \alpha \cdot \beta_\mathrm{est}(z),
    \label{eq:adjust_density}
\end{equation}

For example, two new hazy images with $\alpha =  0.5$ and $\alpha =  2$ are illustrated in Figs.~\ref{haze augmentation}(A) and \ref{haze augmentation}(B), respectively. 
As illustrated in Fig.~\ref{haze augmentation}(C), we can also reverse the location of haze patterns by altering the atmospheric light map $A_\mathrm{est}(z)$ as 
\begin{equation}
    A^{\prime}(z) = 1 - A_\mathrm{est}(z),
    \label{eq:reverse}
\end{equation}
where  $A_\mathrm{est}(z)$ ranges in $[0,1]$. In this case, for those reverse regions that do not contain haze in the original hazy image, we sample $\beta^{\prime}(z)$ to be in $[0.6, 1.25]$,  the range of  $\beta_\mathrm{est}$ in the whole training set. 
Moreover, we can linearly interpolate $A_\mathrm{est}(z)$ and $1-A_\mathrm{est}(z)$ to generate $A^{\prime}(z)$ as 
\begin{equation}
    A^{\prime}(z) = \min (\gamma A_\mathrm{est}(z) + \eta (1 - A_\mathrm{est}(z)), 1),
    \label{eq:interpolate}
\end{equation}
where $\gamma$ and $\eta$ denote the weights for $A_\mathrm{est}(z)$ and $1-A_\mathrm{est}(z)$, respectively.
This allows us to generate complex hazy images, as shown in Fig.~\ref{haze augmentation}(D). In our experiments, we generate additional hazy images for each hazy and clean training pair by adjusting their haze density and location based on the above haze augmentation process.

%% file: 4Experiments.tex
\section{Experiments}
\subsection{Implementation Details}
\textbf{PANet.} For optimizing PANet, we utilize NH-Haze20 dataset~\cite{NH-Haze_2020} that contains non-homogeneous hazy and clean pairs with real-world outdoor scenes. We follow the settings in~\cite{Fu_2021_CVPR,cui2023focal} that utilize $50$ training pairs and $5$ testing pairs in our experiments. During the training process, we employ the Adam optimizer with the initial learning rate of $5 \times 10^{-5}$ that is decayed to $10^{-7}$ by using the cosine annealing strategy to train PANet for $270$ epochs with a batch size of 2. We also randomly crop $256\times256$ patches and utilize random rotation and flipping for augmentation.

\vspace{0.5em}
\noindent\textbf{Dehazing Models.} We adopt three state-of-the-art dehazing models, including DW-GAN~\cite{Fu_2021_CVPR}, DeHamer~\cite{guo2022dehamer}, and FocalNet~\cite{cui2023focal}, to demonstrate the effectiveness of PANet. 
To make a fair comparison, we utilize the $50$ training pairs of NH-Haze20 to train the above-mentioned dehazing models as their baseline following the default training setting in their methods.
To demonstrate the effectiveness of PANet, we utilize PANet to generate additional $400$ training pairs, which is $8$ times larger than the original training set. We combine the augmented $400$ training pairs and the original $50$ training pairs to train the three dehazing models as the enhanced version of their baseline. 
To evaluate the performances of the dehazing models, we choose four real-world hazy image datasets, including NH-Haze20~\cite{NH-Haze_2020} test set, NH-Haze21~\cite{NH-Haze_2021} dataset, O-Haze~\cite{O-HAZE_2018} test set, and I-Haze~\cite{I-HAZE_2018} test set. Specifically, NH-Haze20 test set contains $5$ testing pairs with non-homogeneous haze. Since NH-Haze21 does not provide a test set, we use its training set that contains $25$ pairs collected in non-homogeneous hazy environments for evaluation.
In contrast, O-Haze and I-Haze test sets contain $5$ outdoor and $5$ indoor testing pairs with homogeneous haze, respectively.   


\begin{table}[t!]
\centering
\setlength{\tabcolsep}{0.1mm}
\caption{Quantitative performances of different dehazing methods on NH-Haze20 test set, NH-Haze21 dataset, O-Haze test set, and I-Haze test set. "Baseline" and "+PANet" represent the dehazing performance without and with PANet, respectively.}
\vspace{-0.1in}
\resizebox{\textwidth}{!}{
\begin{tabular}{cc|lr|lr|lr|lr}
\hline\hline
            &
            & \multicolumn{2}{c|}{NH-Haze20~\cite{NH-Haze_2020}} & \multicolumn{2}{c|}{NH-Haze21~\cite{NH-Haze_2021}} & \multicolumn{2}{c|}{O-Haze~\cite{O-HAZE_2018}} &
            \multicolumn{2}{c}{I-Haze~\cite{I-HAZE_2018}}\\
\multicolumn{2}{c|}{Model}     & PSNR (dB) & SSIM & PSNR (dB) & SSIM & PSNR (dB) & SSIM &
PSNR (dB) & SSIM\\ \hline

DW-GAN~\cite{Fu_2021_CVPR} & Baseline  & 21.50 & 0.697  & 18.10  & \bf0.726 & 18.44 & 0.574 & 14.88& 0.403       \\ 
& +PANet  & \bf21.84 (+0.34) & \bf0.704  & \bf18.42 (+0.32)  & 0.708 & \bf20.15 (+1.71) & \bf0.634 & \bf15.47 (+0.59) & \bf0.508       \\ 
\noalign{\hrule height 1.0pt}
DeHamer~\cite{guo2022dehamer} & Baseline  & 20.01 & 0.649  & 16.49  & 0.612 & 20.01 & 0.600 & 15.49 & 0.463       \\ 
& +PANet  & \bf20.73 (+0.72) & \bf0.650  & \bf17.05 (+0.56)  & \bf0.627 & \bf20.64 (+0.63) & \bf0.650 & \bf16.22 (+0.73) & \bf0.563       \\ 
\noalign{\hrule height 1.0pt}
FocalNet~\cite{cui2023focal} & Baseline  & 20.31 & 0.646  & 16.51  & 0.632 & 18.28 & 0.622 & 15.29 & \bf0.417       \\ 
& +PANet  & \bf20.76 (+0.45) & \bf0.682  & \bf17.87 (+1.36)  & \bf0.700 & \bf20.64 (+2.36) & \bf0.639 & \bf15.41 (+0.12) & 0.374       \\ 
\noalign{\hrule height 1.0pt}
\multicolumn{2}{c|}{Average Gain}  & \bf+0.50 & \bf+0.015  & \bf+0.75  & \bf+0.022 & \bf+1.57 & \bf+0.042 & \bf+0.48  & \bf+0.054        \\ 
\hline\hline
\end{tabular}}
\label{Tab:real dehazing}
\vspace{-0.2in}
\end{table}

\subsection{Performance Evaluations}
\textbf{Quantitative Performance Comparison.}  
In Table~\ref{Tab:real dehazing}, we compare the dehazing performances of three baselines and their PANet-enhanced versions, where ``Baseline'' and ``+PANet'' denote the dehazing performance without and with PANet, respectively.
As shown in Table ~\ref{Tab:real dehazing}, The PANet-augmented data significantly improve the average PSNR performances of the three dehazing models, including DW-GAN~\cite{Fu_2021_CVPR}, DeHamer~\cite{guo2022dehamer}, and  FocalNet~\cite{cui2023focal} by $0.50$ dB, $0.75$ dB, $1.57$ dB, and $0.48$ dB on NH-Haze20, NH-Haze21, O-Haze and, I-Haze, respectively. These evaluation results demonstrate that PANet can effectively help boost the performances of  deep dehazing models under various  haze conditions.

\begin{figure*}[t!]
\centering
\includegraphics[width=1\columnwidth]{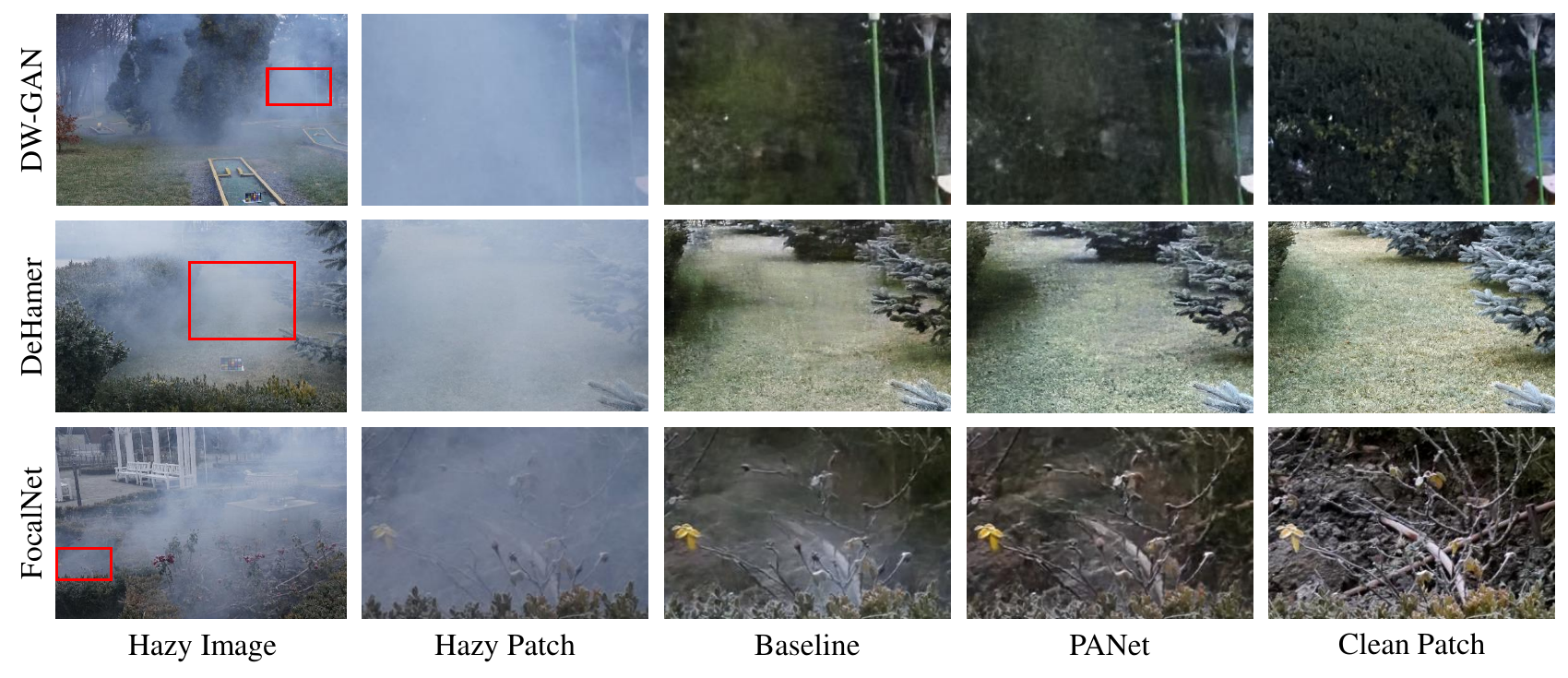}
\vspace{-0.3in}
\caption{Qualitative performance comparison on NH-Haze20~\cite{NH-Haze_2020} testing set.}
\label{fig:VisualizationNH-Haze20}
\vspace{-0.1in}
\end{figure*}

\begin{figure*}[t!]
\centering
\includegraphics[width=1\columnwidth]{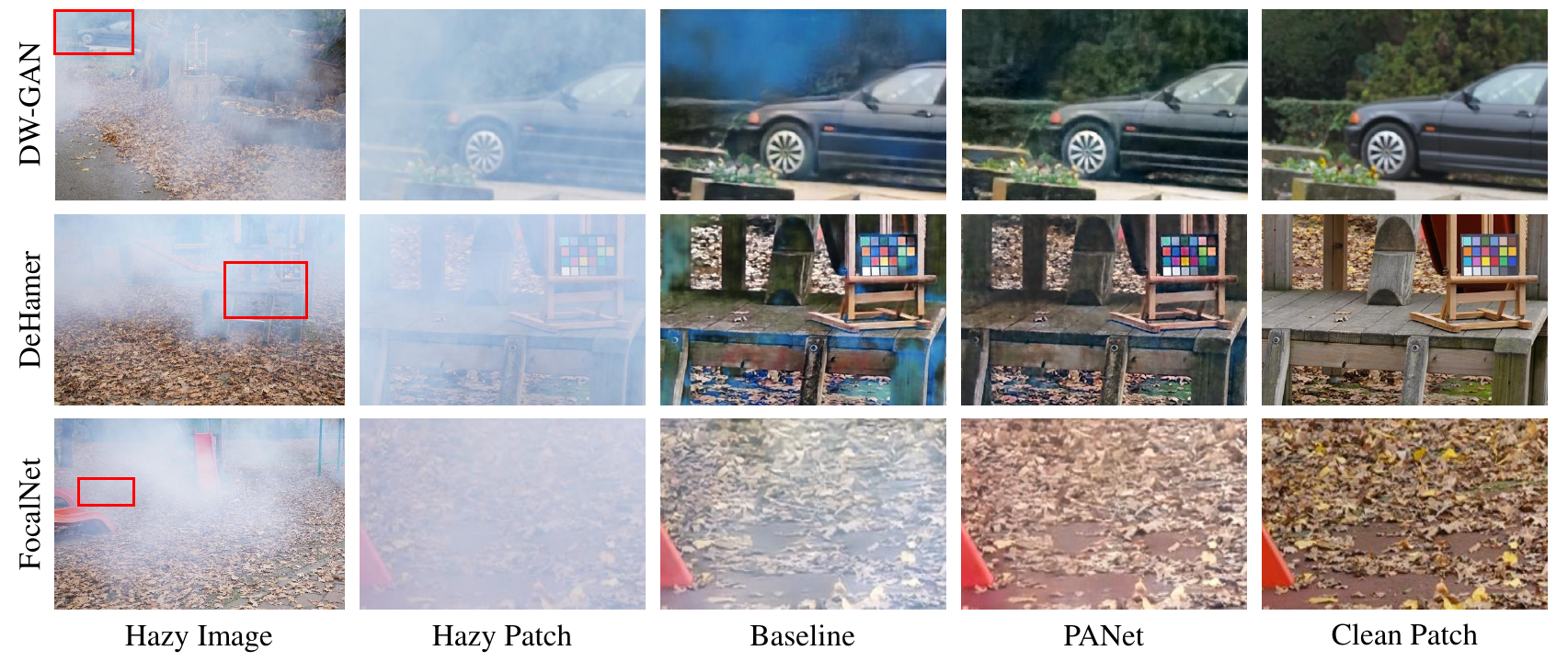}
\vspace{-0.3in}
\caption{Qualitative performance  comparison on NH-Haze21~\cite{NH-Haze_2021} dataset.}
\label{fig:VisualizationNH-Haze21}
\vspace{-0.1in}
\end{figure*}

\begin{figure*}[t!]
\centering
\includegraphics[width=1\columnwidth]{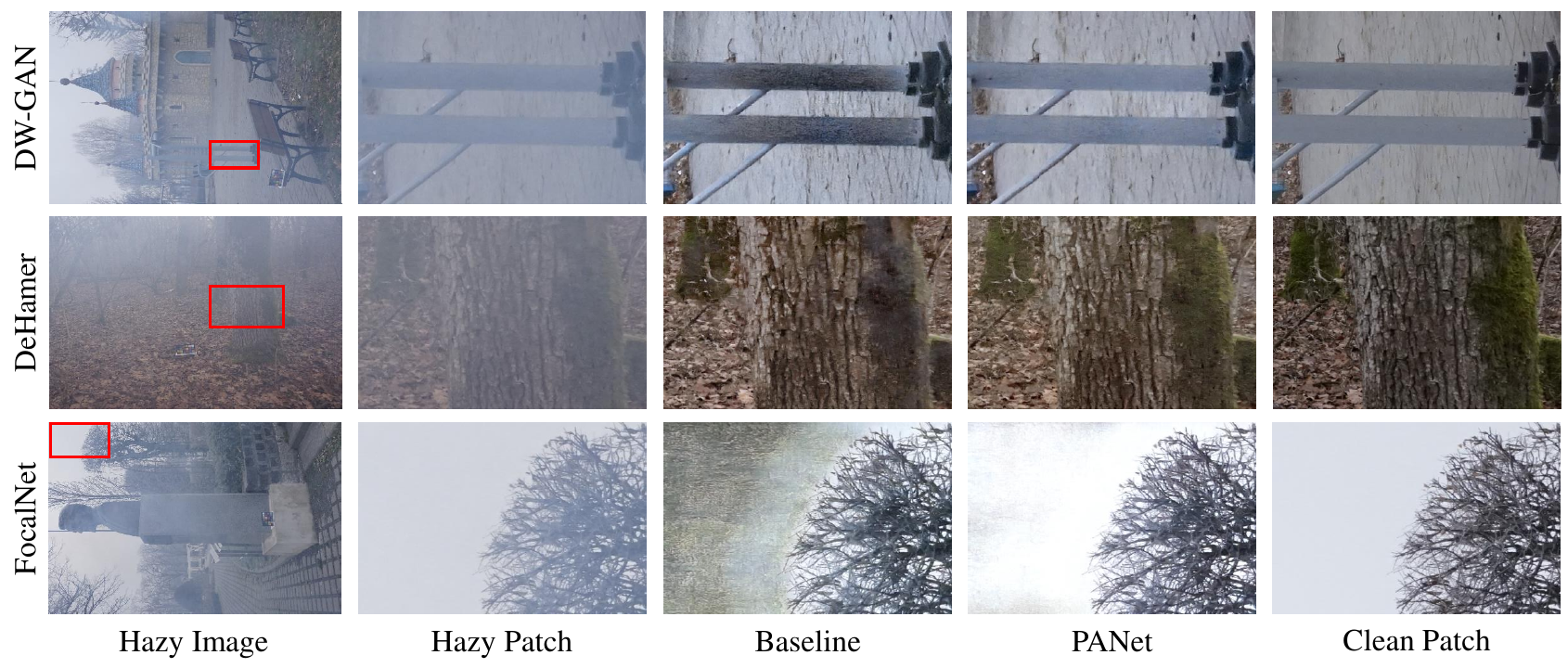}
\vspace{-0.3in}
\caption{Qualitative Comparison on O-Haze~\cite{O-HAZE_2018} testing set}
\label{fig:VisualizationO-Haze}
\vspace{-0.2in}
\end{figure*}

\begin{figure*}[t!]
\centering
\includegraphics[width=1\columnwidth]{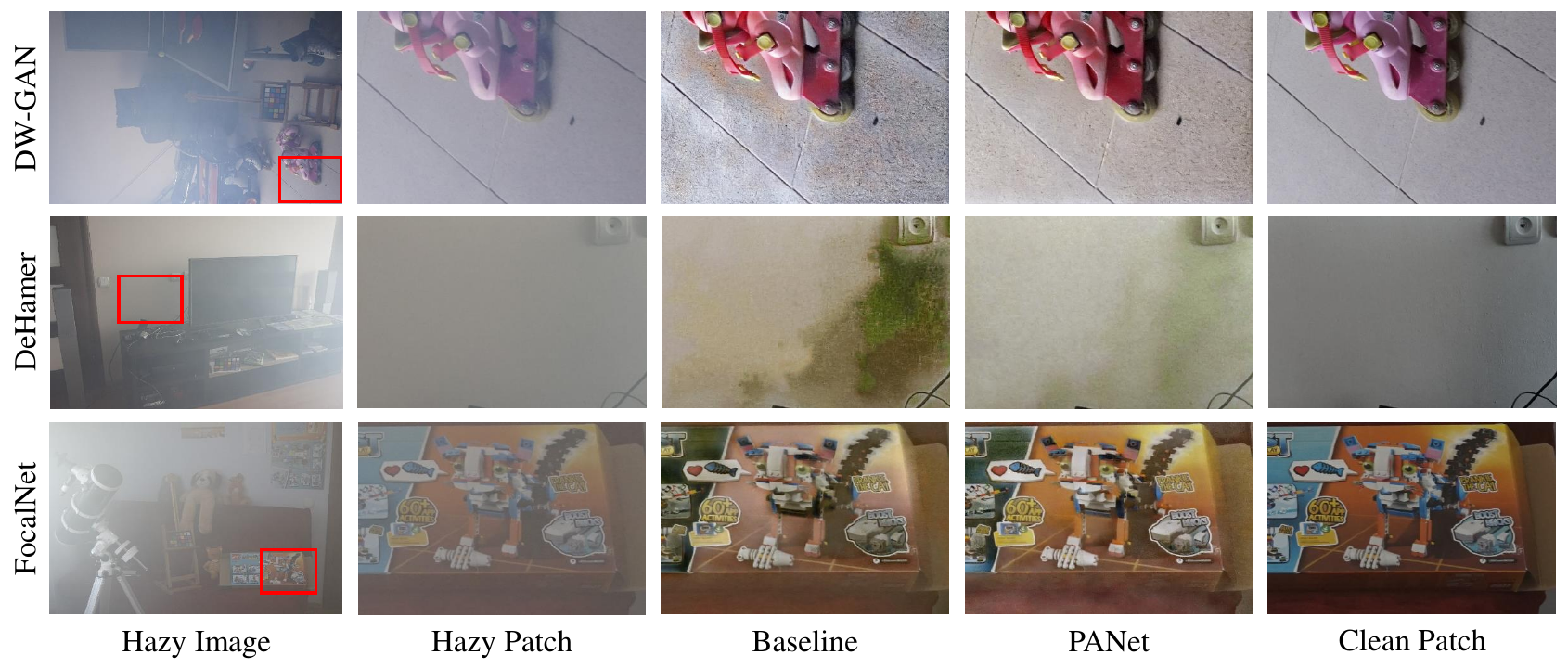}
\vspace{-0.3in}
\caption{Qualitative Comparison on I-Haze~\cite{I-HAZE_2018} testing set}
\label{fig:VisualizationI-Haze}
\vspace{-0.05in}
\end{figure*}

\vspace{0.5em}
\noindent\textbf{Qualitative Performance Comparison.} 
We demonstrate the dehazed results of dehazing models with or without using PANet in Figs.~\ref{fig:VisualizationNH-Haze20},~\ref{fig:VisualizationNH-Haze21},~\ref{fig:VisualizationO-Haze}, and~\ref{fig:VisualizationI-Haze}. 
In Fig.~\ref{fig:VisualizationNH-Haze20}
and Fig.~\ref{fig:VisualizationNH-Haze21}, we visualize some dehazed results on non-homogeneous haze datasets: NH-Haze20 and NH-Haze21. Compared to their baselines, PANet-enhanced models achieve significant visual quality improvements by removing unwanted hazy artifacts or correcting color distortions.
Fig.~\ref{fig:VisualizationO-Haze} and Fig.~\ref{fig:VisualizationI-Haze} visualize some dehazing results on homogeneous haze datasets: O-Haze and I-Haze. Again, the PANet-enhanced dataset can also significantly boost the performances of state-of-the-art models under homogeneous haze conditions in both outdoor and indoor scenarios.
These visuals show that PANet is effective in augmenting both homogeneous and non-homogeneous hazy images in real-world scenarios.

\subsection{Ablation studies}
In the ablation studies, we analyze the impact of PANet on the dehazing performance of FocalNet on NH-Haze20 test set, where``Baseline'' denotes the PSNR performance of FocalNet trained on NH-Haze20 training set without using the additional training pairs augmented by PANet.

\begin{table}[t!]
\centering
\setlength{\tabcolsep}{3.5mm}
\caption{Effect of the Data-driven Haze Refiner (DHR) $N_\mathrm{ref}(\cdot)$ in PANet.}
\vspace{-0.1in}
\begin{tabular}{c|ccc}
\hline
FocalNet \cite{cui2023focal}  & Baseline & w/o DHR & with DHR \\
\hline
NH-Haze20 (PSNR) & 20.31 dB  &  20.22 dB &  \bf20.76 dB\\
\hline\end{tabular}
\label{tab:ablation studyPHM}
\vspace{-0.1in}
\end{table}

\begin{figure*}[t!]
\centering
\includegraphics[width=1\columnwidth]{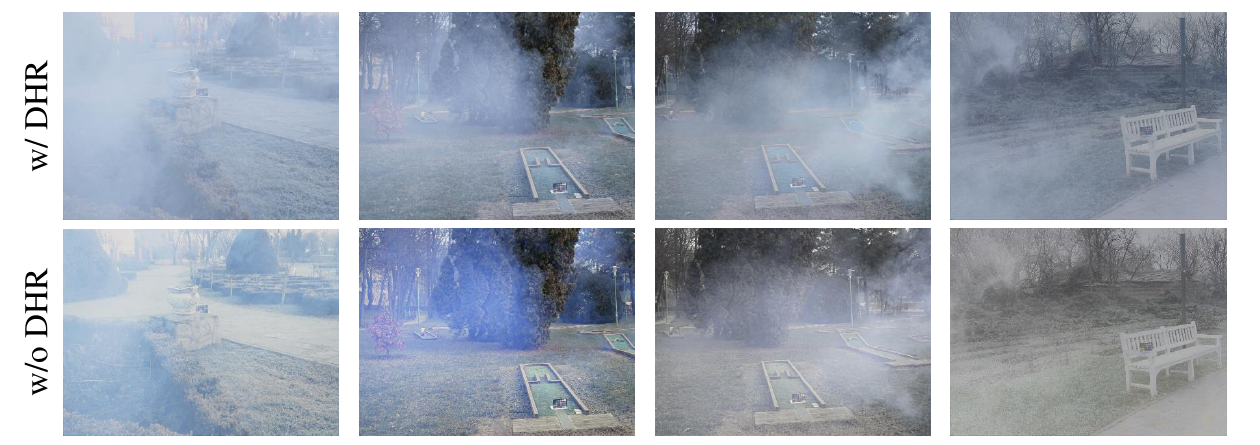}
\vspace{-0.3in}
\caption{Effect of the Data-driven Haze Refiner (DHR) in PANet. First row: Hazy images generated by PANet with DHR. Second row: Hazy images generated by PANet without DHR.}
\label{fig:ablationPHM}
\vspace{-0.05in}
\end{figure*}

\vspace{0.5em}
\noindent\textbf{Effect of the Data-driven Haze Refiner (DHR) $N_\mathrm{DHR}(\cdot)$.} 
To verify the effectiveness of DHR, we compare the dehazing performance enhanced by PANet with or without using DHR. As shown in Table~\ref{tab:ablation studyPHM}, ``w/o DHR'' and ``w/ DHR'' denote the dehazing performances of PANet without and with DHR, respectively. The results show that PANet without DHR, which generates hazy images by solely using the physical scattering model, cannot improve the Baseline, due to the inaccuracy of the estimated haze parameters.
Besides, we compare the qualitative results of the generated hazy images for``w/o DHR'' and ``w/ DHR'' in Fig.~\ref{fig:ablationPHM}. Without using DHR, the generated hazy images contain unrealistic color tones and transparency compared to the ones using DHR, which demonstrates the importance of DHR in PANet.

\vspace{0.5em}
\noindent\textbf{Effect of the Depth Refinement Module (DRM).}
Since the pre-trained depth estimator suffers from a domain gap when addressing unseen clean images, this may degrade the accuracy of the physical scattering model~\cite{othman2022enhanced,lou2023simhaze}. Therefore, we use DRM to refine the initial depth map. We compare the performance with or without using DRM in Table~\ref{tab:ablation studiesDRM}, where 
``w/o DRM'' and ``w/ DRM'' denote PANet without and with DRM, respectively. The results show that PANet without DRM cannot improve the Baseline, which demonstrates the importance of DRM in PANet.

\vspace{0.5em}
\noindent\textbf{Effect of the Number of Augmented Images.}
To analyze the impact of the number of training pairs generated by PANet on the dehazing performance, we use different numbers of training pairs augmented by PANet to improve the Baseline trained on $50$ training pairs.
As shown in Table~\ref{tab:ablation studynumber}, we generate additional $200$ ($400\%$ of the original data size), $400$ ($800\%$ of the original size), and $600$ training pairs ($1,200\%$ of the original size). The results show that the dehazing performance improves with the number of additional training pairs but tend to get saturated when the number of additional pairs reaches $600$. Therefore, we choose to augment $400$ additional training pairs considering the tradeoff between the training time and the performance gain. 

\begin{table}[t]
\centering
\setlength{\tabcolsep}{3.5mm}
\caption{Effectiveness of the Depth Refinement Module (DRM) in PANet.}
\vspace{-0.1in}
\begin{tabular}{c|ccc}
\hline
FocalNet \cite{cui2023focal}  & Baseline & w/o DRM & w/ DRM \\
\hline
NH-Haze20 (PSNR) & 20.31  &  20.31  &  \bf20.76\\
\hline
\end{tabular}
\label{tab:ablation studiesDRM}
\end{table}

\begin{table}[t!]
\centering
\setlength{\tabcolsep}{2mm}
\caption{Effect of the dehazing performance versus the number of augmented pairs by PANet, where the original training set contains $50$ training pairs.}
\vspace{-0.1in}
\begin{tabular}{c|cccc}
\hline
FocalNet~\cite{cui2023focal}  & Baseline (0\%) & 200 (400\%) & 400 (800\%) & 600 (1200\%) \\
\hline
NH-Haze20 (PSNR) & 20.31  &  20.51  &  20.76 & \bf20.83 \\
\hline\end{tabular}
\label{tab:ablation studynumber}
\vspace{-0.1in}
\end{table}

\begin{figure*}[t!]
\centering
\includegraphics[width=1\columnwidth]{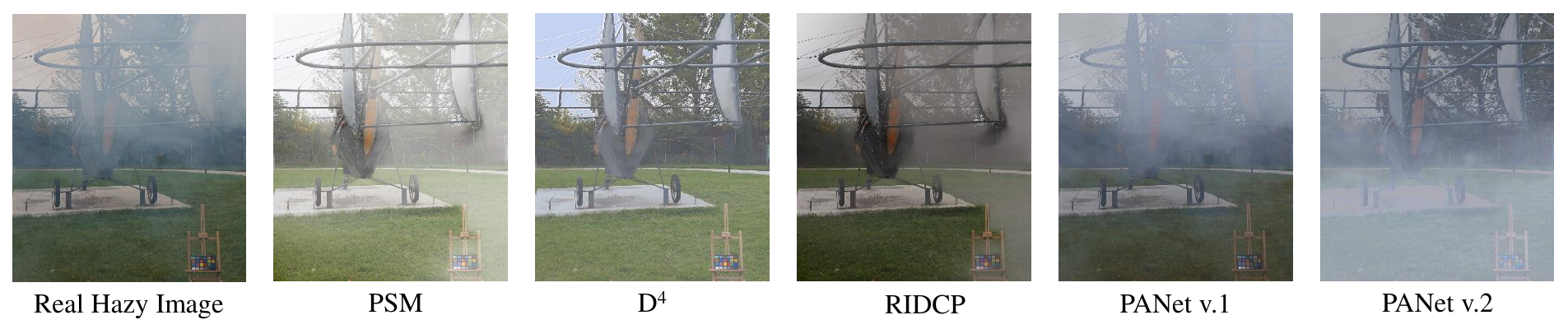}
\vspace{-0.3in}
\caption{Hazy images generated by different augmentation methods, including Physical Scattering Model (PSM), $D^4$~\cite{yang2022self}, RIDCP~\cite{wu2023ridcp}, and PANet.}
\label{fig:differentaugmentation}
\vspace{-0.1in}
\end{figure*}

\begin{figure*}[t!]
\centering
\includegraphics[width=0.9\columnwidth]{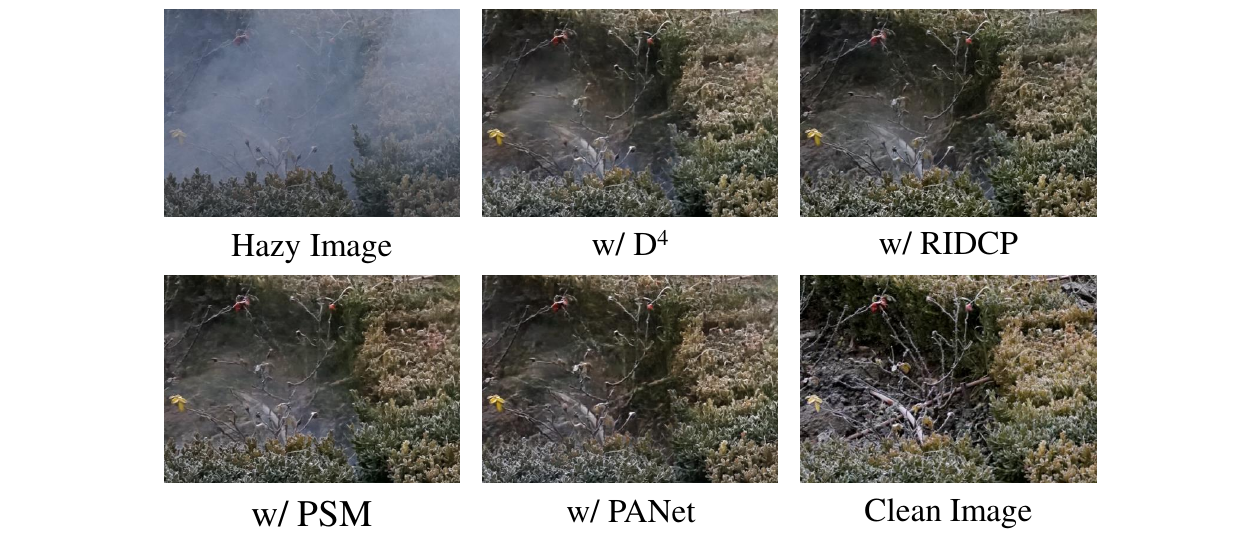}
\vspace{-0.2in}
\caption{Dehazing results with different augmentation methods, including Physical Scattering Model (PSM), $D^4$~\cite{yang2022self}, RIDCP~\cite{wu2023ridcp}, and PANet.}
\label{fig:dehazedifferentaugmentation}
\vspace{-0.05in}
\end{figure*}

\vspace{0.5em}
\noindent\textbf{Comparison with different augmentation methods.}
Finally, we compare PANet with different haze augmentation methods, including uniform haze method: Physical Scattering Model (PSM), GAN-based method: $D^4$~\cite{yang2022self}, and RIDCP~\cite{wu2023ridcp}, where we use these methods to augment the NH-Haze20 training set to improve FocalNet and evaluate on the NH-Haze20 testing set.
As shown in Table~\ref{tab:differentaugmentation}, our PANet achieves the best performance compared to other augmentation methods. In addition, we compare the hazy images generated by these methods in Fig.~\ref{fig:differentaugmentation}.
Among these methods, PSM and RIDCP~\cite{wu2023ridcp} solely rely on the physical scattering models to generate hazy images. Although RIDCP can alter the brightness and color bias, they both cannot generate non-homogeneous hazy images. $D^4$~\cite{yang2022self} applies a cycle-GAN-based architecture to generate hazy images. However, the lack of robustness of GAN increases the difficulty of generating realistic hazy images. In addition, GAN-based methods cannot pixel-wisely control haze conditions to generate diverse hazy images. In contrast, our PANet is a robust network through the physics-guided learning strategy and can pixel-wisely alter hazy conditions to generate diverse non-homogeneous hazy images, which also achieve the best dehazing performances as shown in Fig.~\ref{fig:dehazedifferentaugmentation}.

\begin{table}[t!]
\centering
\setlength{\tabcolsep}{2mm}
\caption{Comparison regarding dehazing performance with different augmentation methods, including Physical Scattering Model (PSM), $D^4$~\cite{yang2022self}, RIDCP~\cite{wu2023ridcp}, and PANet.}
\vspace{-0.05in}
\resizebox{\textwidth}{!}{
\begin{tabular}{c|ccccc}
\hline
FocalNet~\cite{cui2023focal}  & Baseline & w/ PSM & w/ D$^4$~\cite{yang2022self} & w/ RIDCP~\cite{wu2023ridcp} & w/ PANet \\
\hline
NH-Haze20 (PSNR) & 20.31  & 20.41  & 20.26  & 20.57  &  \bf20.76\\
\hline
\end{tabular}}
\label{tab:differentaugmentation}
\vspace{-0.25in}
\end{table}

%% file: 5Conclusion.tex
\section{Conclusion}
In this paper, we proposed a Parametric Augmentation Net (PANet) to generate diverse non-homogeneous hazy images for boosting the performances of dehazing models in real-world scenarios. PANet comprises a Haze-to-Parameter Mapper (HPM) that projects real hazy images into a parameter space and a Parameter-to-Haze Mapper (PHM) that maps them back to hazy images. With PANet, we can pixel-wisely resample the estimated haze parameter maps to generate hazy images with various physically-explainable haze conditions unseen in the training set. This allows us to generate diverse training pairs to improve dehazing performance.  Extensive experimental results demonstrate the high efficacy of PANet in enhancing three state-of-the-art dehazing models on four real-world hazy image datasets.
